\documentclass[review]{elsarticle}

\usepackage{amsmath}
\usepackage{multirow}
\usepackage{algorithm2e}
\usepackage{algorithmicx}
\usepackage{makecell}
\journal{Neurocomputing}

\begin{document}
	
	\begin{frontmatter}
		
		\title{Gram-SLD: Automatic Self-labeling and Detection for Instance Objects}
		
		%% Group authors per affiliation:
		%\author{Rui Wang}
		%\address{School of Instrumentation Science and Opto-electronics Engineering\\
		%	Beihang University\\
		%	Key Laboratory of Precision Opto-mechatronics Technology, China}
		
		%% or include affiliations in footnotes:
		\author[mymainaddress]{Rui Wang}
		\cortext[mycorrespondingauthor]{Corresponding author}
		\ead[email]{wangr@buaa.edu.cn}	
		\author[mymainaddress]{Chengtun Wu}  
		\author[mymainaddress]{Jiawen Xin}	
		\author[mysecondaryaddress]{Liang Zhang}	
		\address[mymainaddress]{School of Instrumentation Science and Opto-electronics Engineering\\Beihang University\\Key Laboratory of Precision Opto-mechatronics Technology, China}
		\address[mysecondaryaddress]{ Department of Electrical and Computer Engineering\\University of Connecticut\\371 Fairfield Way, U-4157, Storrs, Connecticut 06269, United States}

%%%%%%%%%%%%%%%%%%%%%%%%%%%%%%%%%% 摘要 %%%%%%%%%%%%%%%%%%%%%%%%%%%%%%%%
% \begin{abstract}%   <- trailing '%' for backward compatibility of .sty file
% This paper describes the mixtures-of-trees model, a probabilistic 
% model for discrete multidimensional domains.  Mixtures-of-trees 
% generalize the probabilistic trees of \citet{chow:68}
% in a different and complementary direction to that of Bayesian networks.
% We present efficient algorithms for learning mixtures-of-trees 
% models in maximum likelihood and Bayesian frameworks. 
% We also discuss additional efficiencies that can be
% obtained when data are ``sparse,'' and we present data 
% structures and algorithms that exploit such sparseness.
% Experimental results demonstrate the performance of the 
% model for both density estimation and classification. 
% We also discuss the sense in which tree-based classifiers
% perform an implicit form of feature selection, and demonstrate
% a resulting insensitivity to irrelevant attributes.
% \end{abstract}

\begin{abstract}%   <- trailing '%' for backward compatibility of .sty file
Instance object detection plays an important role in intelligent monitoring, visual navigation, human-computer interaction, intelligent services and other fields. Inspired by the great success of Deep Convolutional Neural Network (DCNN), DCNN-based instance object detection has become a promising research topic. To address the problem that DCNN always requires a large-scale annotated dataset to supervise its training while manual annotation is exhausting and time-consuming, we propose a new framework based on co-training called Gram Self-Labeling and Detection (Gram-SLD). The proposed Gram-SLD can automatically annotate a large amount of data with very limited manually labeled key data and achieve competitive performance. In our framework, gram loss is defined and used to construct two fully redundant and independent views and a key sample selection strategy along with an automatic annotating strategy that comprehensively 
consider precision and recall are proposed to generate high quality pseudo-labels. Experiments on the public GMU Kitchen Dataset (\cite{RN13}), Active Vision Dataset (\cite{RN11}) and the self-made BHID-ITEM Dataset (\cite{dataest}) demonstrate that, with only 5\% labeled training data, our Gram-SLD achieves competitive performance in object detection (less than 2\% mAP loss), compared 
with the fully supervised methods. In practical applications with complex and changing environments, the proposed method can satisfy the real-time and accuracy requirements on instance object detection.
\end{abstract}

%%%%%%%%%%%%%%%%%%%%%%%%%%%%%%%%%%%% 关键词 %%%%%%%%%%%%%%%%%%%%%%%%%%%%%%%%%
% \begin{keywords}
%   Bayesian Networks, Mixture Models, Chow-Liu Trees
% \end{keywords}

\end{frontmatter}

%%%%%%%%%%%%%%%%%%%%%%%%%%%%%%%%%% introduction %%%%%%%%%%%%%%%%%%%%%%%%%%%%%%%%%
% \section{Introduction}

% Probabilistic inference has become a core technology in AI,
% largely due to developments in graph-theoretic methods for the 
% representation and manipulation of complex probability 
% distributions~\citep{pearl:88}.  Whether in their guise as 
% directed graphs (Bayesian networks) or as undirected graphs (Markov 
% random fields), \emph{probabilistic graphical models} have a number 
% of virtues as representations of uncertainty and as inference engines.  
% Graphical models allow a separation between qualitative, structural
% aspects of uncertain knowledge and the quantitative, parametric aspects 
% of uncertainty...\\

% {\noindent \em Remainder omitted in this sample. See http://www.jmlr.org/papers/ for full paper.}
\section{Introduction}

Instance object detection refers to recognizing and locating a specific object in an image or a video. It is a core functionality in many applications of computer vision, especially in the humanoid robotics. When intelligent robots work in family or office scenes, we usually require it to have ‘keen eye’ on some specific objects, which often appear in a kind of unstructured environment. In such cases, an instance detection system should tell not only inter-class differences, for example, can vs. box, but also intra-class differences, e.g., soda can vs. coffee can. Thus, how to detect the instances in unstructured environments with complicating factors such as noise, occlusion, random variation in illumination, scales and viewpoints is a big challenge.

In recent years, object detection has become a very active research field thanks to the rapid development of machine learning techniques. Earlier work in this field heavily depends on handcrafted features (\cite{RN8}, \cite{RN159}), which suffer from 
the lack of effective feature extraction and poor generalization ability. The applicationof deep learning (\cite{2020Recent}, \cite{2020Deep}) for object detection(\cite{RN45}, \cite{RN153}, \cite{RN130}, \cite{ZHANG2020180}) has shown promising results in the last few years. Recently, more powerful Deep Convolutional Neural Network (DCNN) models such as R-CNN (\cite{RN45}), Faster RCNN(\cite{RN153}),R-FCN (\cite{RN130}) has become the new frontier of the field and have further pushed the envelope of object detection performance to new records, generally these networks focus on category-level object detection. Due to the similarity of task, it is common to use category-level detection networks to detect instance objects (\cite{RN32}, \cite{RN136}) by transfer learning. DCNN models typically require a large-scale labeled dataset for supervised training. Currently, category detection databases (e.g., PASCAL VOC, COCO) and instance detection databases (e.g., GMU Kitchen Dataset, AVD Dataset) are both open accesses. However, when training the instance object detection network, it is necessary to provide many labeled samples of instance objects under multi-view, multi-scale, and different illumination conditions. Although the public category detection datasets provide many labeled samples of the same category, there are usually only a handful of labeled samples for each instance object, which is far from a sufficient number of examples needed by deep learning networks. At present, the most widely used instance detection databases, GMU Kitchen Dataset and AVD Dataset, are mainly for kitchen and home scenes, including common objects such as soda cans and canned boxes, but this is not enough to meet our actual needs. In practical applications, for new environments and new detection objects, we still need to build instance detection datasets for specific tasks. This is usually a time-consuming and exhausting process. As a result, how to use a small number of labeled samples to achieve high-precision detection becomes the bottleneck in the development of instance target detection algorithms.
In some scenarios, such as the instance object detection mentioned above, computer-assisted medical image analysis, etc., it is difficult to obtain many labeled samples. Therefore, using a large number of unlabeled examples to improve learning performance is a critical research area. The existing mainstream technologies in this area include semi-supervised learning (\cite{RN175}), transductive learning (\cite{RN58}), and active learning (\cite{RN120}). Semi-supervised learning uses a large amount of unlabeled data and a small amount of labeled data for pattern recognition. Compared with transductive learning, it has a stronger generalization performance for unknown samples. Compared with active learning, there is no need to interact with the outside world to obtain query results. In particular, a co-training-base semi-supervised algorithm is less affected by model assumptions and is simple and effective (\cite{RN165}). This implies that co-training may be a promising approach to instance object detection.
Co-training (\cite{RN48}) states that if there are two sufficient and redundant views, the performance of the learner can be improved by unlabeled samples. Its workflow is as follows: 1) Use a small number of manually labeled samples to train the network initially;2) Forward propagate unlabeled samples in the co-training network to generate pseudo-labels and select better quality pseudo-labels to add to the training set, which achieves self-labeling;3) Use the updated training set to retrain the network. Then, iterating the above steps 2 and 3 until the number of unlabeled pictures falls below a certain threshold.Although, theoretically speaking, this method can lead to high-precision instance object detection if properly, it should be noted that, the performance (i.e., detection accuracy) of the co-training process are directly affected by three key factors: 1)The detection accuracy of the initial model (trained by manually labeled samples). Obviously, a better initial model is the most important factor to ensure the accuracy of the final model detection. If a poor initial model is adopted, the generated prediction results are unreliable and the self-labeling strategy, regardless of how smart it is, won’t be able to generate high-quality pseudo-labels; 2) ndependence of the detector. The reason why the co-training works is that the two detectors have two views of weak dependence. Correlated detectors will output two highly similar pseudo-labels for the same input, which makes it impossible for us to tell whether the pseudo-labels are valid or not; 3)The quality of the pseudo-label. If pseudo-labels with low-quality are updated to the training set, the retrained model will reduce the detection accuracy on the test set due to the increase of incorrect labels. At this stage, the research in this area is not deep enough and no good and mature solutions have been obtained.To address these three challenges above, we propose the Gram-SLD for instance object detection, which includes a key sample selection strategy for selecting samples that can represent the sample distribution and contain more information from unlabeled data. The key samples are then manually labeled to train the initial model. Moreover, a co-training network with gram loss is constructed to ensure the independence of two detectors. Finally, a self-labeling strategy that comprehensively considers accuracy and recall is designed to ensure the quality of pseudo-labels.
The rest of this paper is organized as follows. Section 2 briefly reviews related work. Section 3 presents our method in details. Experimental results are reported in Section 4. Conclusions are drawn in Section 5.
%%%%%%%%%%%%%%%%%%%%%%%%%%%%%%%%%%%%%%%%%%%% 相关工作 %%%%%%%%%%%%%%%%%%%%%%%%%%%%%%%%%%%%%%%%

\section{Related Work}
  \subsection{Instance object detection}   
%%%%%%%%%%%%%%%%%%%%%%%%%%%%%%%%%%%%%%%%%%%% 2.1 %%%%%%%%%%%%%%%%%%%%%%%%%%%%%%%%%%%%%%%%
Object detection can be divided into two key subtasks: object recognition and object positioning. Object recognition refers to determining whether there is an object of interest in the picture and identifying what the object is. Object positioning needs to point out the specific coordinate position of the object of interest in the picture. In the past few decades, there has been a large amount of literature worldwide focusing on the research of object detection (\cite{WU20201}). Traditional target detection methods are mostly based on manually extracted features or templates for recognition and positioning, which have several limitations. For example, while features such as SIFT (\cite{RN36}) and SURF (\cite{RN35}) are scale-invariant and robust to image rotation and illumination changes, which make them suitable for objects with rich texture features, it is difficult to extract stable feature points on non-textured objects. With the rise of deep learning, many scholars have proposed to apply deep neural networks to object detection in place of the traditional artificially designed features. In this endeavor, DCNN, which overcomes the limitations of manual extracted feature to a great extent, becomes one of the most promising method. Specifically, R-CNN first uses a selective search algorithm to extract thousands of regions that may contain objects, then compresses these candidate regions to a uniform size, and sends them to the convolutional neural network for feature extraction. In the last layer, features are entered into an SVM classifier to obtain the classification of the candidate region. Fast-RCNN further improves the computational efficiency by convolving the entire image to obtain features and combining the two steps of candidate region classification and border fitting into one. Note that these methods all rely on selective search to generate candidate regions, which may be very time-consuming. To overcome this issue, Faster R-CNN uses RPN to obtain candidate regions, greatly reducing training time and testing time. YOLO (\cite{RN33}) and SSD (\cite{RN40}) directly regress each bounding box through a single convolutional neural network and predict the probability of the corresponding category, leading to improved training and detection speed. He (\cite{RN34}) proposes the Focal Loss to solve the problem of category uneven distribution of single-stage object detection and designs a single-stage object detection network structure based on RetinaNet with the addition of an FPN module. This method achieves both high accuracy and high speed. Although different object detection networks have different structural frameworks, they usually have the same components. First, the backbone of the network is the part that converts the image into a feature map, such as ResNet-50 without the last fully connected layer. Then, RoIExtractor extracts RoI-wise features from single or multiple feature maps using operations such as RoI Pooling. Finally, RoIHead uses RoI as input to perform RoI-wise specific tasks, such as bounding box classification and regression, etc.

The instance object detection problem refers to the detection of a specific instance in a category. This problem can be solved by high performing category object detection models such as the classic Faster R-CNN, RetinaNet. However, DCNN models always require a large-scale labeled dataset for supervised training. How to overcome the challenge of a small number of training samples? The mainstream solution is data expansion (\cite{RN160}, \cite{RN90}) by which a small number of images are manually labeled to obtain a mask of the instance object, and many detection training samples are generated by pasting these instances into a complex background. Cut, Paste and Learn first uses the semantic segmentation network to obtain the mask of the instance object. Then, the images of the instance objects to be pasted are enhanced and border processing, such as blurring, are performed to generate more semantic synthetic samples. Synthesizing proposes a data expansion method that fully considers geometric and semantic information. The background image uses the RANSAC (\cite{RN134}) method to obtain some flat surfaces, and the semantic segmentation is used to obtain counters, tables, etc., combining the two to get the possibility place to paste. These methods have the following two problems. One is the mask. The workload of labeling the mask is much higher than that of the bounding box. Although it is currently proposed to use threshold segmentation or semantic segmentation based on deep learning algorithms, the mask accuracy of the former is not reliable enough and the latter undoubtedly increases the workload. Second, the position of the mask pasted on the background may have semantic errors. These pasting-based methods for instance object detection are limited by the above two reasons and leave unlabeled samples wasted. For these reasons, semi-supervised training (\cite{RN155}), which can use a large number of easily available unlabeled samples to improve the performance of the learner, come into notice. The current mainstream semi-supervised learning algorithms include generative models (\cite{RN155}, \cite{RN176}, \cite{RN149}), semi-supervised SVM (\cite{RN58}), graph-based semi-supervised (\cite{RN128}, \cite{RN77}; \cite{RN126}) and co-training (\cite{RN48}).
  \subsection{Co-training}
Co-training, as one of the most recognized approach in semi-supervised learning, was first proposed in 1998 and was mainly used in natural language processing (NLP) and content-based image retrieval (CBIR). During co-training, if there are two sufficient and redundant views, which can be the attribute set or features of the data, the performance of the learner can be improved by unlabeled samples. For example, an image can be described by the visual information of the image itself, or it can be described by its associated text information. At this time, the attribute set extracted by the visual information forms a view of the image, and the attribute set corresponding to the text information forms another view. In addition, sufficient, in this case, means that any one of these two attribute sets are enough to train a strong learner, while redundant means that each attribute set is conditionally independent of the other. In NLP, Pierce and Cardie (\cite{RN151}) use co-training for phrase recognition and the recognition error rate is reduced by 36\%. Using (\cite{RN60}, \cite{RN61}) co-training for syntactic analysis and performance is also significantly improved. Hwa (\cite{RN62}) propose an active semi-supervised syntactic analysis method based on co-training, which can reduce the amount of manual labeling by half. In CBIR, Zhou (\cite{RN64}, \cite{RN163}) apply co-training to CBIR. Experiments show that this technology can effectively improve the retrieval performance on COBRL image database.

In fact, it is often difficult to obtain independent views. Therefore, some researchers try to design co-training algorithms that do not require strict independent views. It is a key issue in the development of co-training. For classification problem based on co-training, Nigam and Ghani (\cite{RN49}) prove that when the feature set is "sufficiently large", dividing the feature set into two parts as two new views randomly will also achieve good results. Goldman and Zhou (\cite{RN52}) obtain two different base classifiers on the same training set by using different decision trees. Later they use noise learning theory and achieve better performance (\cite{RN67}). However, the method frequently uses cross-validation in learning, which increases training time. Zhou et al. propose Tri-training (\cite{RN164}) and Co-forest (\cite{RN144}). At present, there are few studies on regression problems based on co-training. Among these limited results, Zhu (2005) propose a graph regularization algorithm to solve the semi-supervised regression problem. Zhou and Li (\cite{RN68}) propose COREG to perform semi-supervised regression for co-training. Brefeld (\cite{RN69}) learn from the regularization framework and use prediction differences to improve the views. Abeny (\cite{RN61}) propose to use weak independence to replace the previous strong independence and the result proved that co-training still has an effect. On this basis, Balcan (\cite{RN125}) prove that under the premise that the data distribution satisfies the expansion hypothesis, co-training still works. Wang and Zhou (\cite{RN70}) analyze that if there are two strong learners, co-training still also be effective while meeting a certain degree of weak independence. To solve the independence problem in co-training, existing algorithms either choose to use artificially designed feature sets of high-dimension or use multiple learners. The former requires rich machine learning experience and needs to be redesigned in different environments. The biggest problem of the latter is that it is difficult to adjust the parameters and the structure of the model is complicated. To solve the independence between views, Gatys (\cite{RN135}) propose that the gram matrix of the features represents a way of expressing the input image. They then uses the gram matrix to measure the difference in picture styles and generate pictures of different styles. Below, we introduce the function and principle of the gram matrix.
  \subsection{Gram Matrix}
For a two-dimensional matrix, its gram matrix is defined as
\begin{equation}
	     G=W^TW=\begin{bmatrix} w_1^Tw_1 & w_1^Tw_2 & \cdots  & w_1^Tw_N\\
		w_2^Tw_1 & w_2^Tw_2 & \cdots  & w_2^Tw_N \\
		\vdots & \vdots & \ddots  & \vdots\\
		w_N^Tw_1 & w_N^Tw_2 & \cdots  & w_N^Tw_N
	\end{bmatrix} =
	\begin{bmatrix} g_{11} & g_{12} & \cdots  & g_{1N}\\
		g_{21} & g_{22} & \cdots  & g_{2N} \\
		\vdots & \vdots & \ddots  & \vdots\\
		g_{N1} & g_{N2} & \cdots  & g_{NN}\end{bmatrix},
\end{equation}
where $w_i\in R^{M*l}$ is the i-th column vector of the matrix W, $g_{ij}$ is the dot product of $w_i$ and $w_j$, representing a combination of the i-th column vector and the j-th column vector of the matrix W. $g_{ij}>0$ indicates that the two vectors are positively correlated. $g_{ij}<0$ indicates that the two vectors are negatively correlated. $g_{ij}=0$ indicates that the two vectors are not correlated. The magnitude of $|g_{ij}|$ indicates the correlation between these two vectors. For an orthogonal base W, since the column vectors are all pairwise orthogonal, that is, the dot product is 0, the diagonal elements of the gram matrix are all 1, and the non-diagonal elements are all 0. For a non-orthogonal base, the non-diagonal element of its gram matrix is non-zero. Each element in the gram matrix represents a combination of two columns of the original matrix.

Inspired by Gatys et al., we believe that using the gram matrix to obtain the weakly dependent views is a solution that can realize the self-labeling and high-precision detection of instance objects combined with co-training. There are, of course, another two prerequisites that co-training also depends on, the accuracy of the initial model and the quality of the pseudo-labels. We will discuss each of these in the following sections.

%%%%%%%%%%%%%%%%%%%%%%%%%%%%%%%%%%%%%%%%%%%%%%% 提出方法 %%%%%%%%%%%%%%%%%%%%%%%%%%%%%%%%%%%%%%%%%%%
\section{Proposed Approach}

The goal of our work is to develop a powerful co-training framework called Gram Self-labeling and Detecion (Gram-SLD) for instance object detection with limited labeled data and massive unlabeled data, as shown in Figure 1. In this section, we first introduce the overall network structure and training process of the algorithm and then describe the solutions to solve the three key challenges of co-training.
\begin{figure}[h]  %htbp
    \centering
    \includegraphics[width = .8\textwidth]{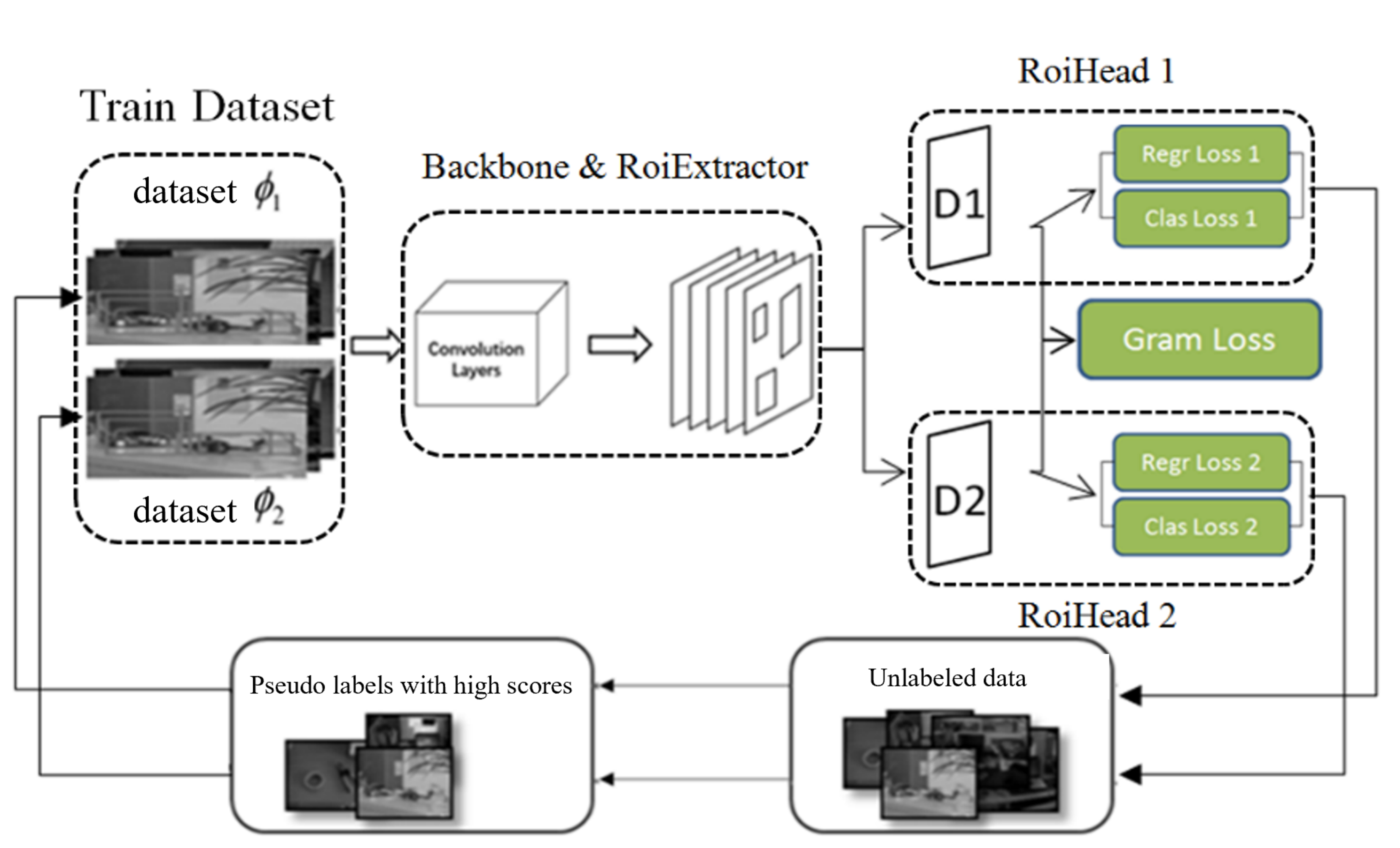}
    \caption{Overview of our Gram-SLD framework for instance object detection.}
    %\label{fig:myphoto}
\end{figure}

\noindent \textbf{The structure of Gram-SLD:} 
Note that Gram-SLD is a general co-training framework based on gram loss, rather than a specific network structure. Therefore, we can apply it to any mainstream category object detection network (referred to as the original network), such as RetinaNet, Faster R-CNN, etc., and make the resulting network suitable for instance object detection tasks with few labeled samples. In this framework, the output of RoiExtractor of the original network uses two 3*3 convolution layers to obtain two features called D1 and D2. These two features are connected to their respective RoiHeads, while the two branches of detector 1 and detector 2 share the same Backbone and RoiExtractor. The loss during training consists of two parts: one is the detection loss including location regression loss and recognition classification loss, the other is the gram loss proposed to solve the independence problem in co-training. This will be introduced in detail in Subsection 3.2. It should also be noted that the original network needs to have a relatively high detection performance for the proposed method to succeed. Our numerical experiments suggest that the test accuracy of original network needs to be above 70\% in public database VOC.

\noindent \textbf{The training process of Gram-SLD:}  
The overall work flow of the proposed Gram-SLD is as follows:1) Select a small number of key samples from unlabeled samples for manual annotation according to the key sample selection strategy and input them into the training set $\phi_1$ of the detector 1 and the training set $\phi_2$ of the detector 2 respectively;2)Based on the selected original detection network, construct Gram-SLD and use the initial key samples to train the Gram-SLD;3) Forward the remaining unlabeled samples in Gram-SLD to obtain two prediction results. According to self-labeling strategy selects samples with high score pseudo label and updates them to the training set;4) Use the updated training set to retrain Gram-SLD. Repeat steps 3 and 4 until the number of unlabeled pictures is less than threshold; then stop the training process.

\noindent \textbf{Solutions to three key challenges in co-training:} 1) Key sample selection strategy to solve the challenge of accuracy of the initial model: Cluster unlabeled samples, select the samples with the largest entropy in each cluster as the key samples and manually label key samples for training. Experiments show that the initial model trained with key samples has high detection accuracy and good generalization ability. Details of the key sample selection strategy are given in Subsection 3.1; 2) Gram loss to solve the challenge of detector independence: This loss function represents the view difference between detector 1 and detector 2. As the gradient drops, the gram loss decreases and the view difference increases, indicating that the view independence is satisfied. This will be introduced in Subsection 3.2; 3)Self-labeling strategy to solve the challenge of the quality of the pseudo-labels: A scoring method that comprehensively considers accuracy and recall is proposed. Through this self-labeling strategy, high-quality pseudo-labels can be obtained. The self-labeling strategy will be introduced in Subsection 3.3.

%%%%%%%%%%%%%%%%%%%%%%%%%%%%%%%%%%%%%%%%%%%%%%%%%%%% 3.1 key sample selection %%%%%%%%%%%%%%%%%%%%%%%%%%%%%

\boldmath
\subsection{\textbf{Sample Selection Strategy}}
\unboldmath
\noindent\textbf{Transfer learning:} 
A well trained detection model before co-training is a foundation of our proposed framework. In other words, before running the co-training framework, we need to pre-train a CNN for feature extraction based on a large object detection dataset, e.g., ImageNet. Since several detection networks are already available in the literature (e.g., RetinaNet and Faster RCNN) and have achieved remarkable success in visual recognition, our Gram-SLD adopts transfer learning, i.e., directly employs the original detection networks and their pre-trained models in ImageNet as initial parameters. Given the initially selected training set of manually annotated samples, we further fine-tune the Gram-SLD to learn discriminative feature representations. However, a small number of arbitrarily selected samples may be infeasible to supervise the training of such deep learning models for object detection. To address this problem, a key sample selection procedure is proposed next to provide big initial performance boost for detection model.

\noindent \textbf{Key sample selection:}
The co-training method is based on the similarity within the training data and the generalization ability of neural network to carry out semi-supervised learning. The key sample selection process is to identify representative training data in the first place for manual annotation. Apparently, since the initial network’s generalization ability is limited, the key samples for initial training should represent the distribution of the overall samples. Otherwise, it would be difficult for Gram-SLD to update unlabeled samples that are very different from the key samples during self-labeling. As co-training progresses, the newly labeled samples will become more and more biased towards the distribution of the initial samples. This will eventually make it difficult for unlabeled training samples to be fully utilized, and the error of deviation will gradually accumulate. In addition, in the same detection scene, the key samples should contain more information, so that the initially trained Gram-SLD learns features of complex samples. This will enhance the prediction capability for simple samples to be used later. Therefore, to address these two challenges, we propose below a novel data mining method called key sample selection strategy, which can automatically select the representative key samples for manual labeling in the training set.
 
In our proposed strategy, the HSV color histogram is first calculated for each sample. Specifically, the H, S, and V channels are quantified using unequal intervals in accordance with human color perception. The H channel is divided into 8 bins, while the S and V channels are each divided into 4 bins. We then integrate the three-dimensional histogram into a one-dimensional vector with 128 bins, which is compatible with Euclidian distance calculation. Next, the training samples are clustered into K classes in terms of the 128 bin HSV color histogram. Note that since hierarchical clustering (\cite{RN161}) can avoid the initial value sensitivity problem encountered by K-means (\cite{RN16}), we apply hierarchical clustering to the HSV description-based clustering task above. Moreover, we choose Calinski-Harabaz Index (CH) (\cite{RN158}) as an indicator to measure the quality of clustering and determine the best number of clusters based on CH. For each class thus obtained, the top M images with the largest entropy are selected as the key samples. Here, the entropy represents average information content of image data defined by
\begin{equation}
	 E=-\sum_{j=0}^{255}p_jlogp_j,
\end{equation}
where $p_j$ represents the proportion of pixels in the image with a gray value of j. Consequently, a total of K*M samples are selected as the initial key sample data to be used for manual labeling. With these K*M key labeled samples, we can train the Gram-SLD.

%%%%%%%%%%%%%%%%%%%%%%%%%%%%%%%%%%%%%%%%%%%%%%%%%%%% 3.2 Gram Loss %%%%%%%%%%%%%%%%%%%%%%%%%%%%%
\subsection{Gram Loss}

As mentioned above, Gram-SLD is a neural network that combines co-training with Gram Loss. Unlabeled samples are annotated by using two detectors during co-training, and successful co-training requires independence of these two detectors. To address this issue, gram loss is adopted in our work to ensure that the correlation of the two detectors decreases during the gradient descent process and guarantees independence. To achieve this, we first divide the mainstream category object detection networks into two kinds, convolutional-kind and fully connected-kind. For both kinds of networks, we first define a unified form of gram matrix and then introduce gram loss based on gram matrix at the end of this subsection.

\noindent \textbf{Gram matrix in CNN:} 
The feature map of ROI obtained by RoIExtractor in different convolutional neural networks is not always the same. We referred to the networks, which have the feature map of ROI as a one-dimensional vector, as fully connected-kind networks (e.g., Faster R-CNN). On the other hand, we referred to the networks, which has the feature map of ROI as a three-dimensional tensor, as convolutional-kind networks (e.g., RetinaNet). Note that the gram matrix used in linear algebra is defined for two-dimensional matrices. In a convolutional neural network, however, the feature map may be a three-dimensional tensor or a one-dimensional vector. This calls for a generalization of the gram matrix to be applicable to the CNN features. Specifically, if the feature map is a three-dimensional tensor, each feature is a two-dimensional matrix, and the elements of the gram matrix represent a combination of these matrices. If the feature map is a one-dimensional vector, then each feature is a scalar, and the elements of the gram matrix should represent the combination of scalars. To formalize the expression, we can replace the “dot product” used in the definition of gram matrix in linear algebra by matrix-based Frobenius inner product in the case of three-dimensional tensor feature map (i.e., convolutional-kind networks) and by simple scalar multiplication in the case of one-dimensional vector feature map (i.e., fully connected-kind networks). No matter what form the feature map is, the elements of the gram matrix is the combination of features, which represents an expression of the feature map to the input image. The definition of the generalized gram matrix is shown in pseudo code and illustrated in Figure 2.
\begin{figure}[h]  %htbp
    \centering
    \includegraphics[width = .8\textwidth]{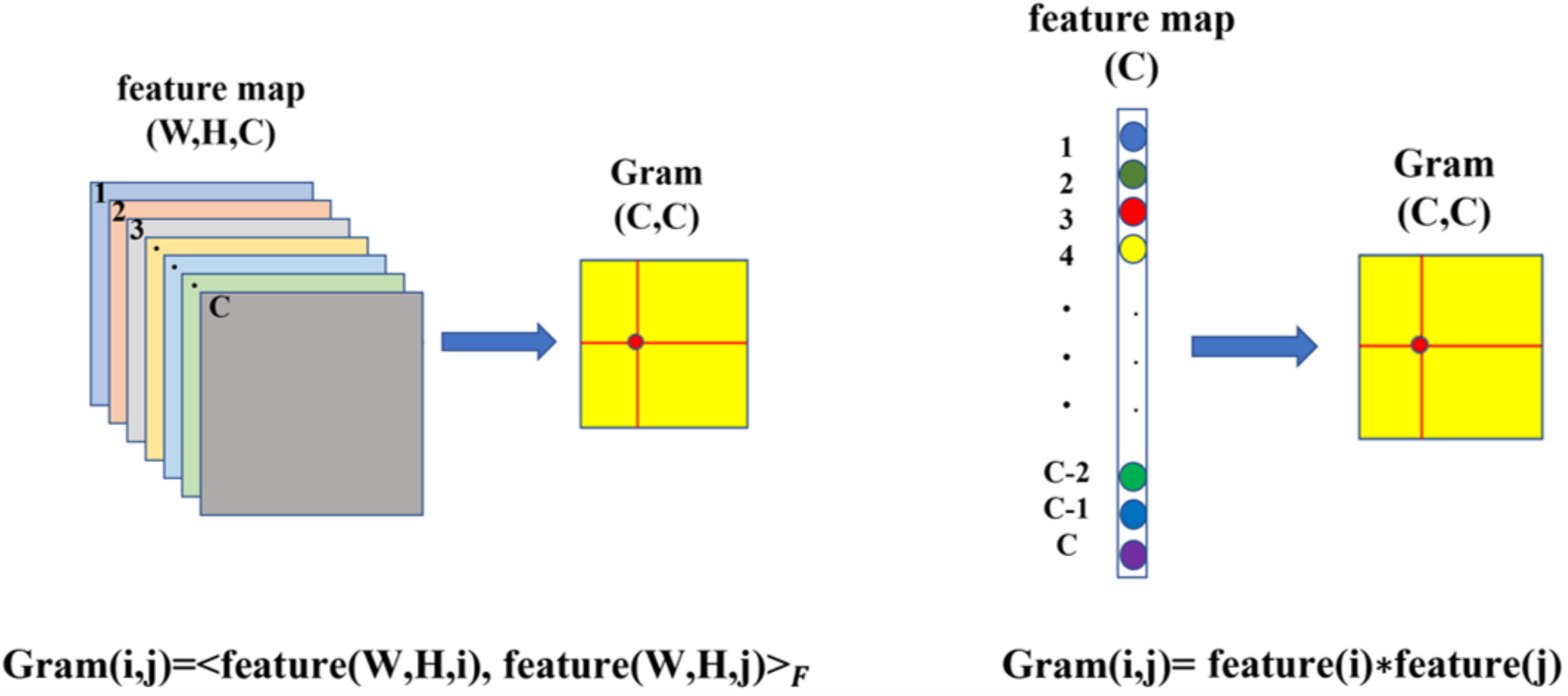}
    \caption{Gram matrix of feature map of different forms.}
    %\label{fig:myphoto}
\end{figure}
%%%%%%%%%%%%%%% fake code %%%%%%%%%%%%%
\RestyleAlgo{ruled}
\begin{algorithm}[h] 

  \caption{Identify Gram matrix}
  
  \KwData{Feature, LayerType}
  
  \KwResult{Gram Matrix}
  
  $C \gets Feature.shape[-1]$
  
  \For{$i = 1$; $i<C$; $i++$ }
   {
    \For{$j = 1$; $j<C$; $j++$ }
    {
     \uIf{$NetType == Convoluntinal$}
     {
      $Feature.shape=(M,N,C)$
      
      $feature1 \gets Feature(M,N,i)$
      
      $feature2 \gets Feature(M,N,j)$
      
      $Gram(i,j) \gets <feature1,feature2>_F$
     }
     \ElseIf{$NetType == fully connected$}
     {
      $Feature.shape=(C)$
      
      $feature1 \gets Feature(i)$
     
       $feature2 \gets Feature(j)$
       
      $Gram(i,j) \gets feature1*feature2$
     }
    }
 }
\end{algorithm}

\noindent \textbf{Gram loss:} 
As mentioned above, the output of RoiExtractor of original network uses two 3*3 convolution layers to obtain two features, D1 and D2, which are connected to their respective RoiHeads. The two resulting branches, D1 and D2, share Backbone and RoiExtractor. Regarding the choice of the original detection network, note that Zhou proposes that “when two fully redundant views do satisfy independence, through co-training, the accuracy of the weak learner (in the case of two classification problems, the accuracy is slightly higher than 50\%) trained by unlabeled samples can be increased to any height.” The scenario considered in this paper is instance object detection. At present, the detection accuracy of a detector on a public category database such as VOC2007 is generally used as an evaluation index. As a preliminary study, we have tried to use a low-precision object detector such as YOLO v1 (the detection accuracy of which on VOC2007 is 63.4\%) and improve it using Gram-SLD. However, it is observed that while the detection accuracy increases after the first iteration of co-training, continuing iterative training leads the accuracy to drop instead of further increasing. This is probably due to the low initial accuracy of the model, resulting in low quality of pseudo-label and negatively affecting the operation of co-training. Therefore, we claim that an original network, which achieves at least close to or higher than 70\% mAP on VOC2007 test dataset, can be adopted for Gram-SLD improvement. Based on this standard, detectors, such as Faster R-CNN (VGG16), RetinaNet (ResNet50), etc., can be used as original network in Gram-SLD. During training, the total loss can be defined as
\begin{equation}
    total\_loss=loss_{D1}+loss_{D2}+gram\_loss*\alpha.
\end{equation}
Among them, gram\_loss is defined as the reciprocal of the mean square error of the Gram matrix of D1 and D2:
\begin{equation}
	gram\_loss=1/Mse(Gram1,gram2),
\end{equation}
In addition, $loss_{D1}$ is the sum of the classification loss and regression loss of the detector D1 branch defined as
\begin{equation}
	loss_{D1}=L_{cls\_1}+L_{loc\_1},
\end{equation}
while $loss_{D2}$ is defined similarly by the above formula. The classification loss $L_{cls}$ and regression loss $L_{loc}$ of different networks are not the same. For detail, please refer to related paper. Note that $loss_{D1}+loss_{D2}$, the first part of the total loss, implies the detection accuracy of the two detectors, while gram\_loss, the second part of the total loss, is the indicator of the difference between the two detectors D1 and D2, that is, the independence requirements for co-training. The coefficient $\alpha$ is to avoid a large difference between the loss terms. During the training process, gram loss gradually decreases by the gradient descent. Figure 3 shows the heat map of the gram matrix difference between the two features (detectors) before and after co-training. In the heat map, darker the color of a point in the heat map, the closer the values of corresponding entry in the gram matrices of the two detectors. Similarly, brighter color implies that the difference between the gram matrices of the two detectors is large. This figure clearly shows that the difference of gram matrix of the two detectors increases significantly before and after co-training, and, thus, the independence of the views of the two detectors is greatly enhanced.
\begin{figure}[h]  %htbp
    \centering
    \includegraphics[width = .8\textwidth]{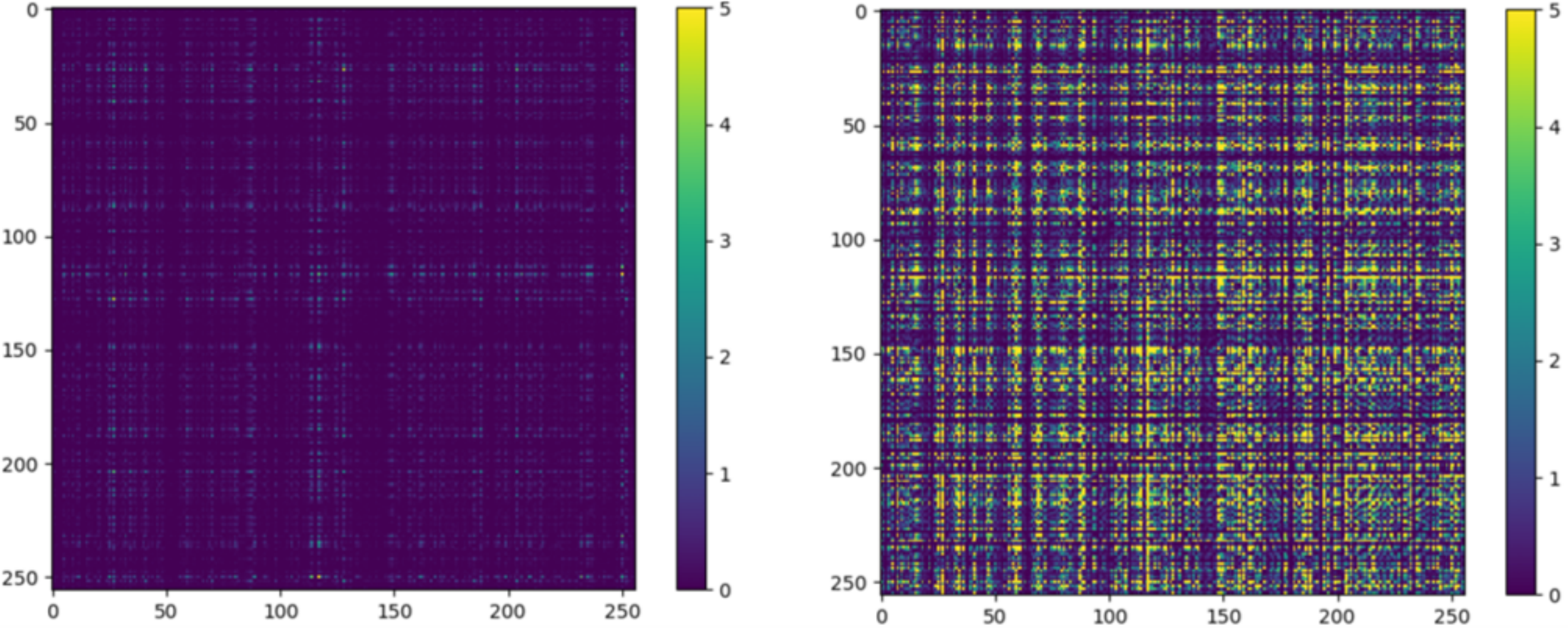}
    \caption{difference of gram matrix before and after co-training.}
    %\label{fig:myphoto}
\end{figure}

%%%%%%%%%%%%%%%%%%%%%%%%%%%%%%%%%%%%%%%%%%%%%%%%%%%% 3.3 Self-labeling Strategy %%%%%%%%%%%%%%%%%%%%%%%%%%%%%
\subsection{Self-labeling Strategy}

The process of self-labeling should try to avoid errors introduced by pseudo labels, that is, to ensure the quality of the updated sample labels, which can guide the detection model to learn in the correct direction. The so-called high quality updated samples, specifically in the object detection task, is the samples with pseudo-labels of high precision and high recall. These samples will be added to the training pool. Note that the result of detector forward propagation contains the recognition confidence of each bounding box, while the update of the sample is based on the entire picture. Therefore, to ensure the high quality of the picture, the confidence of each bounding box and the recall of bounding box in the picture needs to be considered. If we do not consider the recall of a single picture, there will be many missed bounding boxes. In a picture, if only one bounding box of high confidence is detected, the algorithm adds the entire picture to the training pool. Then, the object that is difficult to detect in the picture is conveniently regarded as the background. This means the more difficult the object is to detect, the more difficult it is to update to the training pool. At this time, the training imbalance may appear. To avoid this problem as much as possible, we propose a self-labeling strategy, which comprehensively considers precision and recall.

Specifically, when using the learned Gram-SLD to predict in unlabeled pool, for each unlabeled sample, detector 1 generates pseudo label A1 and detector 2 generates pseudo label A2. The pseudo label includes the candidate bounding boxes and the confidence of each bounding box. Based on this, we score an unlabeled sample x based on the following scoring rules:
\begin{equation}
	score(x)=\sum_iI_{acc1(w_i)>\delta_{acc}} * I_{acc2(w_i)>\delta_{acc}} * I_{iou(w_i)>\delta_{iou}},
\end{equation}
where $w_i$ is the target object that exists in both pseudo-labels A1 and A2, $acc1(w_i)$ and $acc2(w_i)$ are the confidence of the bounding boxes of $w_i$ in A1 and A2, respectively, $iou(w_i)$ is the IoU of those two bounding boxes, and $I_A$ is the indicator function:
\begin{equation}
	 I_A=\begin{cases}
		1,\quad if\, A\, is\, true, \\
		0,\quad if\, A\, is\, false.
	\end{cases} 
\end{equation}
In other words, the score of sample x is the total number of bounding boxes that meet all three criteria (referred to as valid bounding box). Note that under this definition, the precision is accounted for by the confidence threshold $\delta_{acc}$ and IoU threshold $\delta_{iou}$, while the number of valid bounding box measures the recall of the entire sample. Next, we calculate the score of all samples and select those with scores over a threshold into the training pool. Note that the samples selected into the training pool should not only be of high quality, but also of contribute to improving the network detection accuracy to a desired extent. Clearly, if the score threshold is too high, the number of updated samples will be small, resulting in the model accuracy not being able to gain improvement. On the other hand, if the score threshold is too low, the quality of the updated samples may suffer. To achieve a balance between these two aspects, the score threshold for cluster k (see Subsection 3.1 for details about the clustering) is set as
\begin{equation}
    \sigma_k=\frac{\beta}{N_k}\sum_{i=1}^{N_k}score(x_i^k),
\end{equation}
where $N_k$ denotes the number of remaining samples in cluster k, $x_i^k$ is the i-th remaining sample in the cluster, and $\beta$ is a parameter that determines the number of samples to be added into the training pool after each iteration. The newly labeled samples with scores greater than the threshold in each cluster is updated to the training pool, while the unselected samples remained be viewed as unlabeled. This process is repeated until the number of unlabeled samples falls below a certain number. Then, it is considered that the remaining unlabeled samples are not enough to improve the performance of the entire network and terminates co-training.
 
%%%%%%%%%%%%%%%%%%%%%%%%%%%%%%%%%%%%%%%%%%%%%%% 实验 %%%%%%%%%%%%%%%%%%%%%%%%%%%%%%%%%%%%%%%%%%%
\section{Experiments}

In this section, we choose two public instance datasets and one self-made dataset. GMU Kitchen Dataset  and AVD Dataset are two widely used public datasets in kitchen and home scenarios. BHID-ITEM Dataset is made to verify the effectiveness of the proposed algorithm in a real office environment. All experiments are carried out using pytorch on NVIDIA-GTX-2080. We set up three sets of experiments in total, namely basic experiment, extended experiment and comparison experiment, each of which is discussed below.

%%%%%%%%%%%%%%%%%%%%%%%%%%%%%%%%%%%%%%%%%%%%%%%%%%%% 4.1 Basic Experiment %%%%%%%%%%%%%%%%%%%%%%%%%%%%%
\subsection{Basic Experiment}
In basic experiment, we verify the effectiveness and universality of the proposed Gram-SLD on three instance detection datasets as mentioned above. We choose both RetinaNet (with ResNet50 being the feature extraction module) and Faster R-CNN (with VGG16 being the feature extraction module) as original detection networks to perform Gram-SLD on each dataset as organized below. It should be noted that in this subsection, the number of key samples and $\beta$  , which determines the number of self-labeled samples, are selected as 5\% and 1.0, respectively. The effects of two parameters will be discussed in Subsection 4.2.
%%%%%%%%%%%%%%%%%%%%%%%% GMU %%%%%%%%%%%%%%%%%%%%%%%%%
\subsubsection{GMU Kitchen Dataset}
\noindent \textbf{Dataset Introduction:} 
The GMU Kitchen dataset includes 9 RGB videos of kitchen scenes for a total of 6728 samples. In this dataset, we target on 11 objects present in the dataset overlapping with the BigBird dataset (\cite{RN12}) objects. The train-test split follows the division of the dataset into three different folds as (\cite{RN13}). The training set contains 6 scenes with a total of 3851 samples, and the test set contains 3 scenes with a total of 2877 samples (default fold: 1st).

\noindent \textbf{Experimental Setup:} 
To conduct co-training, 199 key samples are selected from the total 3851 samples in the training set using the strategy described in Subsection 3.1. Specifically, the 3851 training samples are first divided into 15 clusters based on hierarchical cluster and then the top 5\% of samples with the highest entropy from each cluster are selected as the key samples. The 199 samples are then labeled, while the remaining 3652 samples are set as unlabeled. To study the efficacy of the proposed Gram-SLD framework, especially the effectiveness of self-labeling, the performance of three learning strategies is investigated:1)Only use the 199 labeled samples (i.e., 5\% of all data in the training set) to train the original detection network. This strategy is called Initial Train (IT). In other words, IT only uses the small number of manually labeled samples and does not use the unlabeled ones at all;2) Apply the proposed Gram-SLD framework to improve the original detection network, use the 199 labeled samples for initial training and then the remaining 3652 unlabeled samples for co-training. Clearly, with co-training, this strategy (denoted as Gram-SLD) utilizes the entire training dataset without extra manual labeling;3) First manually label all 3851 samples in the training dataset and then use them to train the original detection network. This strategy is referred to as Fully Supervised (FS).The performance of the strategies above are tested on 2877 samples of the GMU test set. We calculate the difference between the results of Gram-SLD and FS as DIFF. Clearly, DIFF is an indicator of the performance gap between learning based on co-training obtained pseudo-labels (efficient but may be inaccurate) and manually annotated labels (accurate but time-consuming).

\noindent \textbf{Results:} 
Figure 4 shows the mAP curves of the two detectors of Gram-SLD with RetinaNet as the original detection network for seven iterations of co-training. The green line shows the mAP after IT at 0.798, while the red line shows the test accuracy of FS training at 0.847. These two strategies use 5\% and 100\% manually annotated samples, respectively. On the other hand, the proposed Gram-SLD strategy reaches a test accuracy of 0.838 after seven iterations. Compared with IT, this method effectively uses the unlabeled samples (95\% of all samples in the training set) through co-training and improves the mAP by 4\%. Compared with FS, the drop off in mAP of FS is less than 1\%. Gram-SLD requires six iterations to achieve the highest mAP. It consumes a lot of time for multiple rounds of training in exchange for high mAP. Due to the design of the self-labeling strategy in Subsection 3.3, in the first co-training iteration, the number of updated samples is great. So is the improvement of mAP. As the iteration progresses, the number of updated samples decreases and the improvement of mAP becomes less obvious or even fluctuate. Table 1 summarizes the results of the above-mentioned experiments on the RetinaNet and also those for Faster R-CNN. It can be found that the DIFF metric between the proposed method (Gram-SLD) and FS is within 2\%, which implies that Gram-SLD effectively uses unlabeled samples to achieves high performance.
\begin{figure}[t]
    \centering
    \includegraphics[width = .8\textwidth]{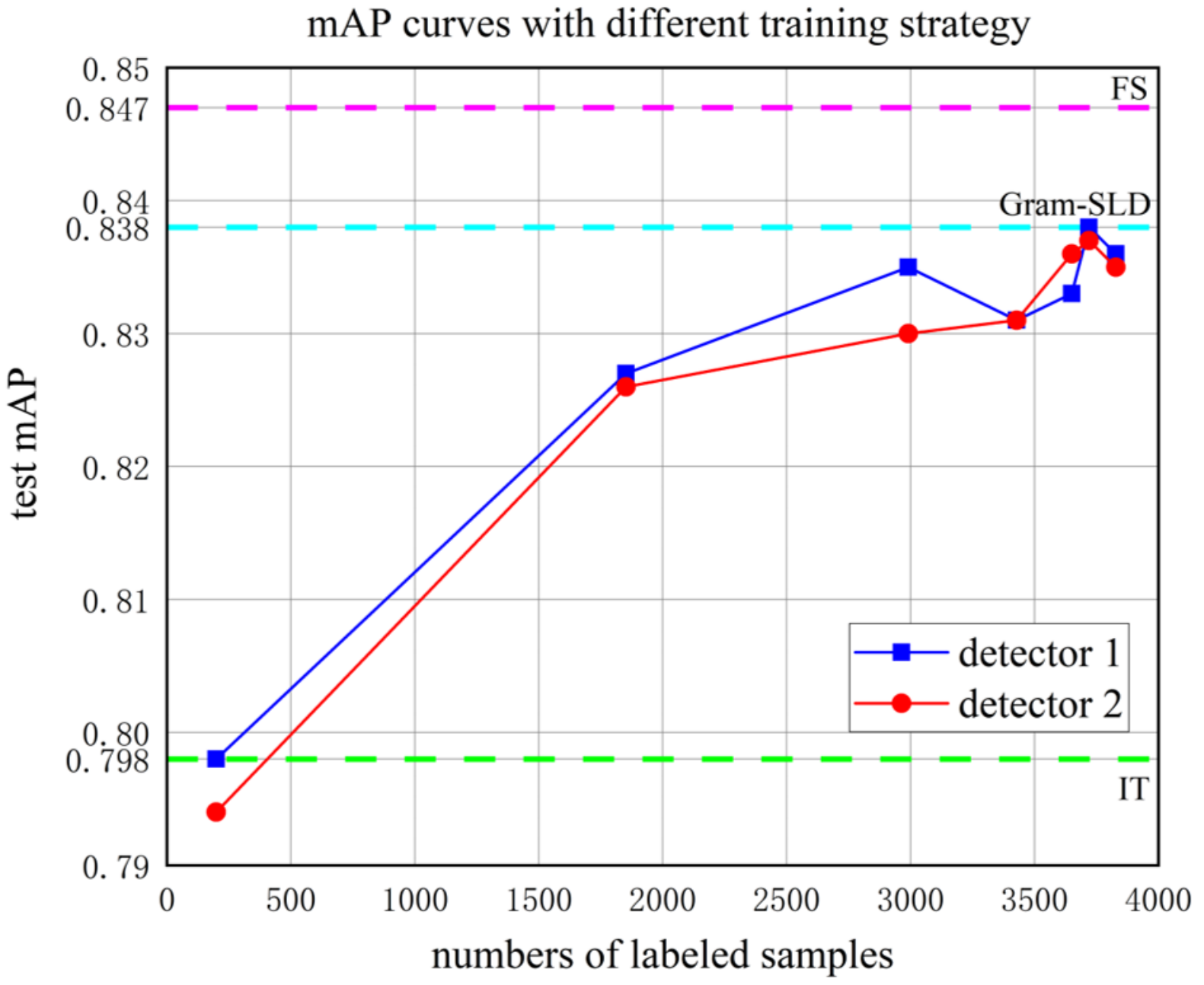}
    \caption{Gram-SLD iterative training mAP curve on RetinaNet \& GMU.}
    %\label{fig:myphoto}
\end{figure}
\begin{table}[h]
    \centering
    \caption{Instance detection accuracy (mAP) on GMU dataset.}
    \begin{tabular}{|c|c|c|c|c|}  % 居左，居中，居右
        \hline
        model & IT & Gram-SLD(proposed) & FS & DIFF\\
        \hline
        RetinaNet & 79.8\% & 83.8\% & 84.7\% & 0.9\%\\
        \hline
        Faster R-CNN & 74.4\% & 79.2\% & 80.8\% & 1.6\%\\
        \hline
    \end{tabular}
\end{table}
%%%%%%%%%%%%%%%%%%%%%%%% AVD %%%%%%%%%%%%%%%%%%%%%%%%%
\subsubsection{AVD Dataset}
\noindent\textbf{Dataset Introduction:} 
AVD contains 17556 samples in 15 scenes and 33 instance objects are annotated. To facilitate the experiment, we choose 7262 samples with 6 common instance objects just like (\cite{RN32}). The official website of the AVD database has a training set and a test set division method. Under the division method, the detection mAP on RetinaNet is only 43.6\%. In other papers using this dataset, (\cite{RN11}) achieves 39\% using SSD, (\cite{RN122}) achieves 41.9\% via Faster R-CNN. Such accuracy certainly cannot meet the requirement of the proposed method as mentioned in chapter 3.2. Therefore, so we re-divide the AVD training set and test set as follows: From the 7262 images, 70\% of the samples (5083) are randomly selected as the training set, and the remaining 30\% (2179) are used as the test set. Compare to the division method of the AVD data set in official website, this method reduces the difficulty of detection, since the pictures in the test set and those in the training set may come from the same video sequence.

\noindent \textbf{Experimental Setup:} 
Similar to the experiment on the GMU database, 255 labeled samples (5\%) are selected from the 5083 samples in the training set, leaving the remaining 4828 samples as unlabeled. The same three learning strategies are tested again.

\noindent \textbf{Results:} 
Table 2 shows the results of the experiments on the RetinaNet and Faster R-CNN. Clearly, the proposed method (Gram-SLD) has a large improvement over IT, that is, it can effectively use unlabeled samples to improve the detection performance.
\begin{table}[h]
    \centering
    \caption{Instance detection accuracy (mAP) on AVD dataset.}
    \begin{tabular}{|c|c|c|c|c|}
        \hline
        model & IT & Gram-SLD(proposed) & FS & DIFF\\
        \hline
        RetinaNet & 74.5\% & 80.8\% & 82.1\% & 1.3\%\\
        \hline
        Faster R-CNN & 68.9\% & 78.1\% & 82.5\% & 4.4\%\\
        \hline
    \end{tabular}
\end{table}

It should be noted that, compared with FS, the effectiveness of the Gram-SLD strategy is better (smaller DIFF metrics) on the GMU dataset (DIFF = 0.9\% and 1.6\% for RetinaNet and Faster R-CNN, respectively) than on the AVD dataset (DIFF = 1.3\% and 4.4\% for RetinaNet and Faster R-CNN, respectively). Both GMU Kitchen Dataset and AVD Dataset are indoor instance object detection datasets, and the amount of data is roughly the same in thousands of pieces. To understand the cause of the performance difference on these datasets, we note that the distributions of small objects in the two datasets are quite different. In the literature, small objects are defined as those: 1)m with width and height below 1/10 of the original image;2) within 32*32 in the COCO public dataset. Since the resolution of the GMU and AVD datasets pictures is 1920*1080, we define a object with area less than 1\% of the original image as a small object, more than 10\% as a big object, and a normal object if the area falls between 1\% and 10\% of the picture. We count the number of small, normal, and big objects in the GMU and AVD datasets (see Figure 5).It is obvious that there are significantly more small objects in the AVD datasets, which increases the difficulty of object detection, and, thus, results in lower detection performance of Gram-SLD.
\begin{figure}[t]
    \centering
    \includegraphics[width = .6\textwidth]{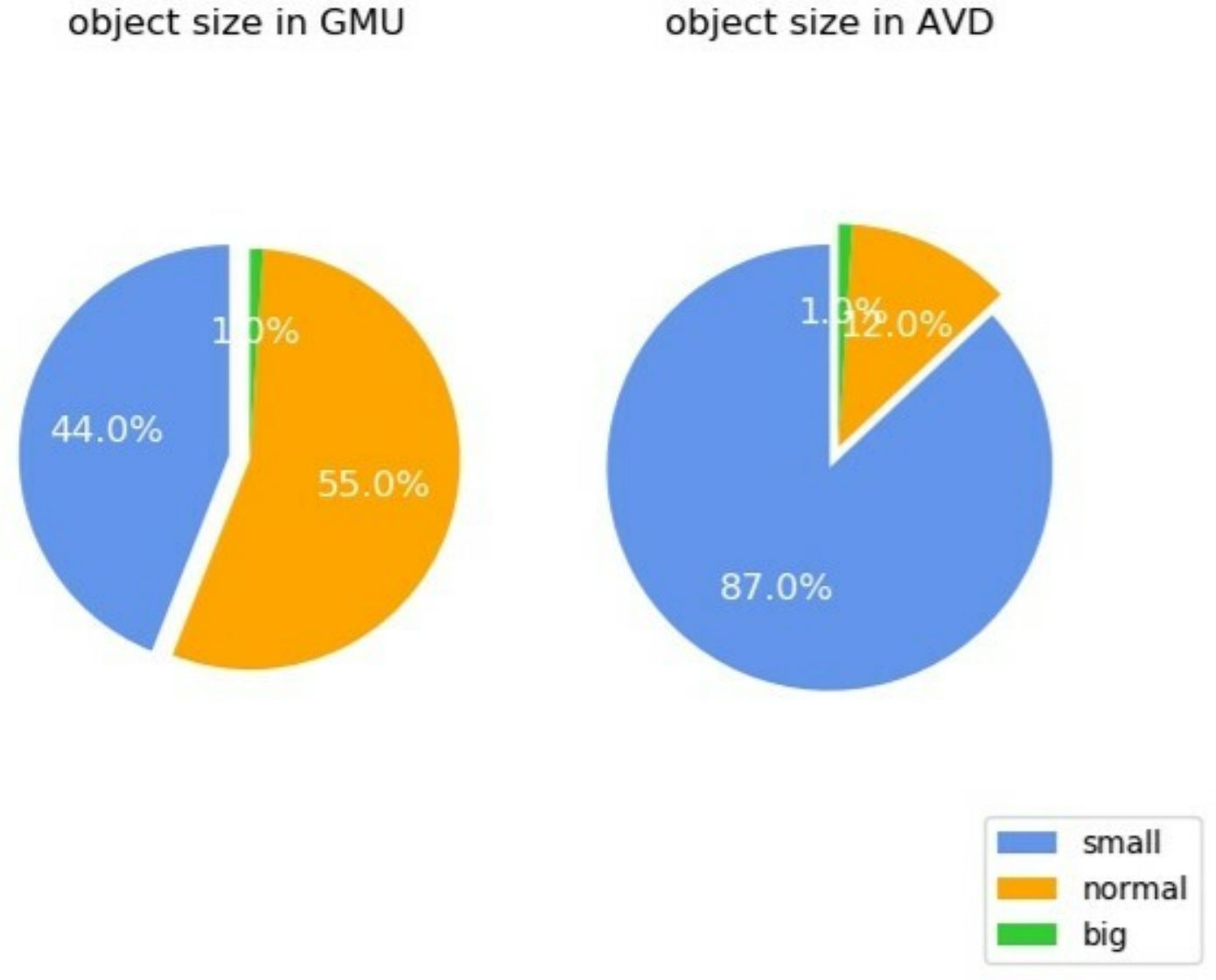}
    \caption{Distribution of object size in GMU and AVD datasets.}
    %\label{fig:myphoto}
\end{figure}
%%%%%%%%%%%%%%%%%%%%%%%% bhid %%%%%%%%%%%%%%%%%%%%%%%%%
\subsubsection{BHID-ITEM Dataset}
\noindent\textbf{Dataset Introduction:}

\noindent \textbf{Experimental Setup:} 
Similar to the experiments on GMU and AVD datasets, 5\% of all samples, i.e., 55 images, are selected from the total 1110 samples in the training set as key samples. The three learning strategies are conducted, with RetinaNet and Faster R-CNN as the original detection network, and their performance is evaluated based on the 611 samples of the BHID-ITEM test set.

\noindent \textbf{Results:} 
The experimental results are summarized in Table 3. It shows that in a more complex indoor scene, the proposed Gram-SLD method improves the detection accuracy over IT through co-training and achieves a similar detection accuracy to FS learning. These observations are consistent with those obtained in the experiments on GMU and AVD datasets.
\begin{table}[h]
    \centering
    \caption{Instance detection accuracy (mAP) on BHID-ITEM dataset.}
    \begin{tabular}{|c|c|c|c|c|}
        \hline
        model & IT & Gram-SLD(proposed) & FS & DIFF\\
        \hline
        RetinaNet & 76.7\% & 83.6\% & 85.5\% & 1.9\%\\
        \hline
        Faster R-CNN & 72.1\% & 81.5\% & 83.2\% & 1.7\%\\
        \hline
    \end{tabular}
\end{table}
%%%%%%%%%%%%%%%%%%%%%%%% sum %%%%%%%%%%%%%%%%%%%%%%%%%
\subsubsection{Summary}
To better visualize the experiment results of this subsection, Figure 6 shows the detection accuracy (mAP) of the IT and the improvements gained by Gram-SLD (over IT) and FS (over Gram-SLD) in a bar chart for all three datasets and two original detection networks tested. As one can see from the figure, the improvement obtained by implementing Gram-SLD (orange section of each bar) is significant, this implies that Gram-S can effectively label and utilize the unlabeled samples through co-training. On the other hand, the accuracy increase by FS over Gram-SLD (green section of each bar, denoted as DIFF in Tables 1-3) is marginal (less than 2\% in all but one cases), implying that the efficacy of Gram-SLD (based on only 5\% labeled training samples) is already close to FS (which relies on a 100\% manually labeled training set). For the only one case where DIFF is greater (4.4\% in Faster R-CNN on AVD), it can be attributed to the fact that the results of IT in this experiment is low (less than 70\%). Note that in this case, Gram-SLD is still capable of improving the detection accuracy by about 10\% over IT, the largest improvement among all cases considered. Thus, based on this set of experiments, we claim that the generality of the proposed method, Gram-SLD, is verified.
\begin{figure}[t]
    \centering
    \includegraphics[width = .6\textwidth]{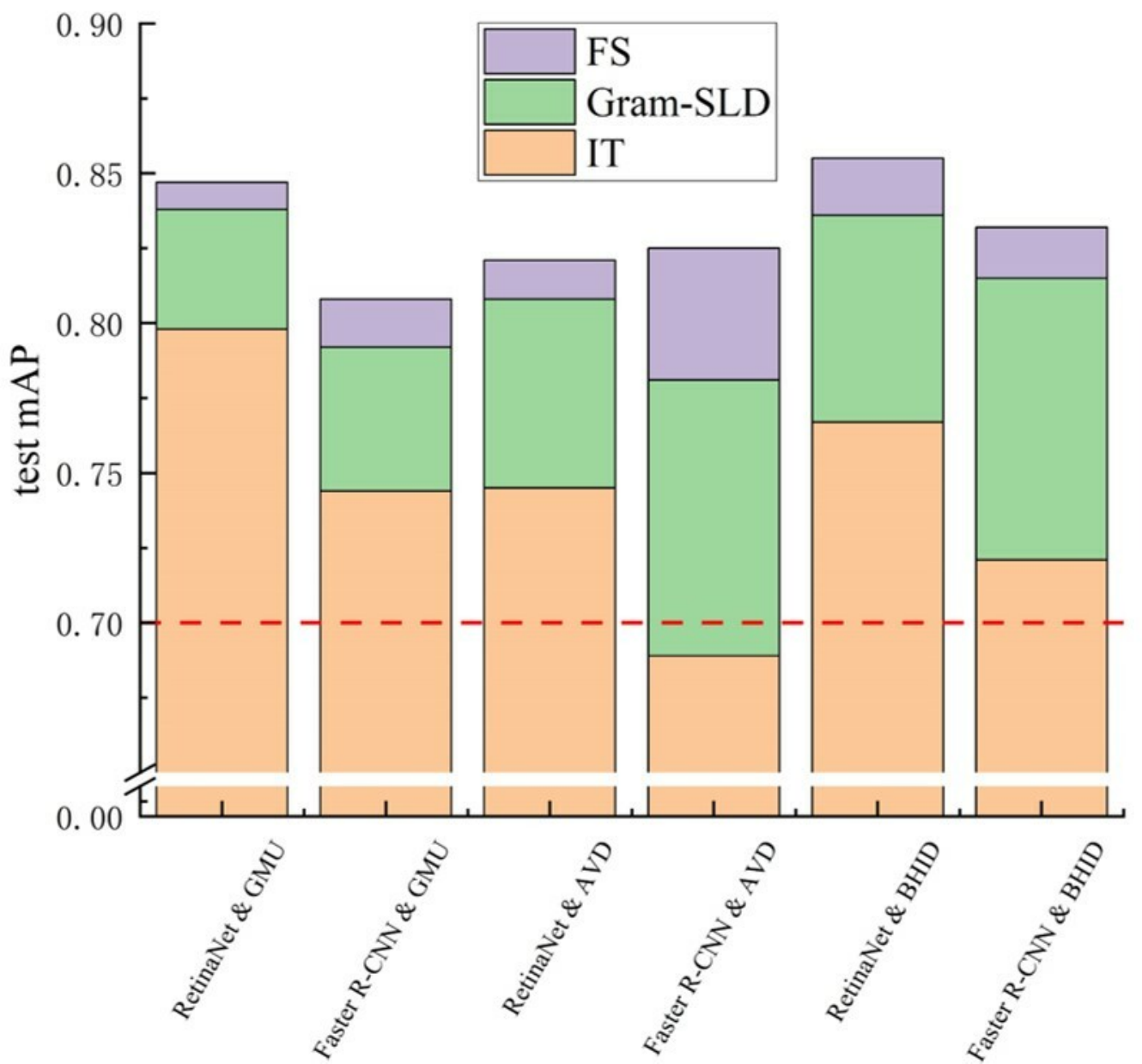}
    \caption{Results of basic experiments.}
    %\label{fig:myphoto}
\end{figure}

%%%%%%%%%%%%%%%%%%%%%%%%%%%%%%%%%%%%%%%%%%%%%%%%%%%% 4.2 Extended Experiments %%%%%%%%%%%%%%%%%%%%%%%%%%%%%
\subsection{Extended Experiments}

In this subsection, we set up three experiments to explore the best parameters of Gram-SLD and the necessity of gram loss. 1) Gram-SLD vs. number of key samples: We train the Gram-SLD with 2\%, 5\%, and 10\% of all training samples selected as initial key samples. Clearly, more initial key samples lead to high detection accuracy of Gram-SLD but at the cost of extra work needed by manual annotation. 2) Gram-SLD vs. parameter $\beta$: We train the Gram-SLD with different values of $\beta$, which determines the number of self-labeled samples to be updated into the training pool in each iteration (see equation (2)). The goal is to explore the effect of the proportion of self-labeled samples on the final detection accuracy and total training time. 3) An ablation experiment on the gram loss to study the effect of the loss on the independence of the two detectors in co-training.
%%%%%%%%%%%%%%%%%%%%%%%% key samples %%%%%%%%%%%%%%%%%%%%%%%%%
\subsubsection{Effects of the numbers of key samples}
This experiment explores the effect of the number of key samples on the performance of Gram-SLD. In this experiment, we use GMU Dataset and choose RetinaNet as the original detection network. Following the key sample selection strategy described Subsection 3.1, we select 2\% (82), 5\% (199), 10\% (396) of the images from the training set as key samples and have them manually labeled. The resulting detection accuracy (mAP) under these three cases are shown in Figure 7. As one can see, when the ratio of key samples is increased from 2\% to 5\%, the detection mAP of Gram-SLD increases from 82.1\% to 83.8\%. However, adding more labeled samples (from 5\% to 10\%) into the key samples only leads to very little improvement (from 83.8\% to 84.0\%) as co-training completes. Consequently, we conclude that using 5\% labeled data as the initial key samples is necessary and sufficient for our Gram-SLD method to achieve satisfactory performance.
\begin{figure}[t]
    \centering
    \includegraphics[width = .6\textwidth]{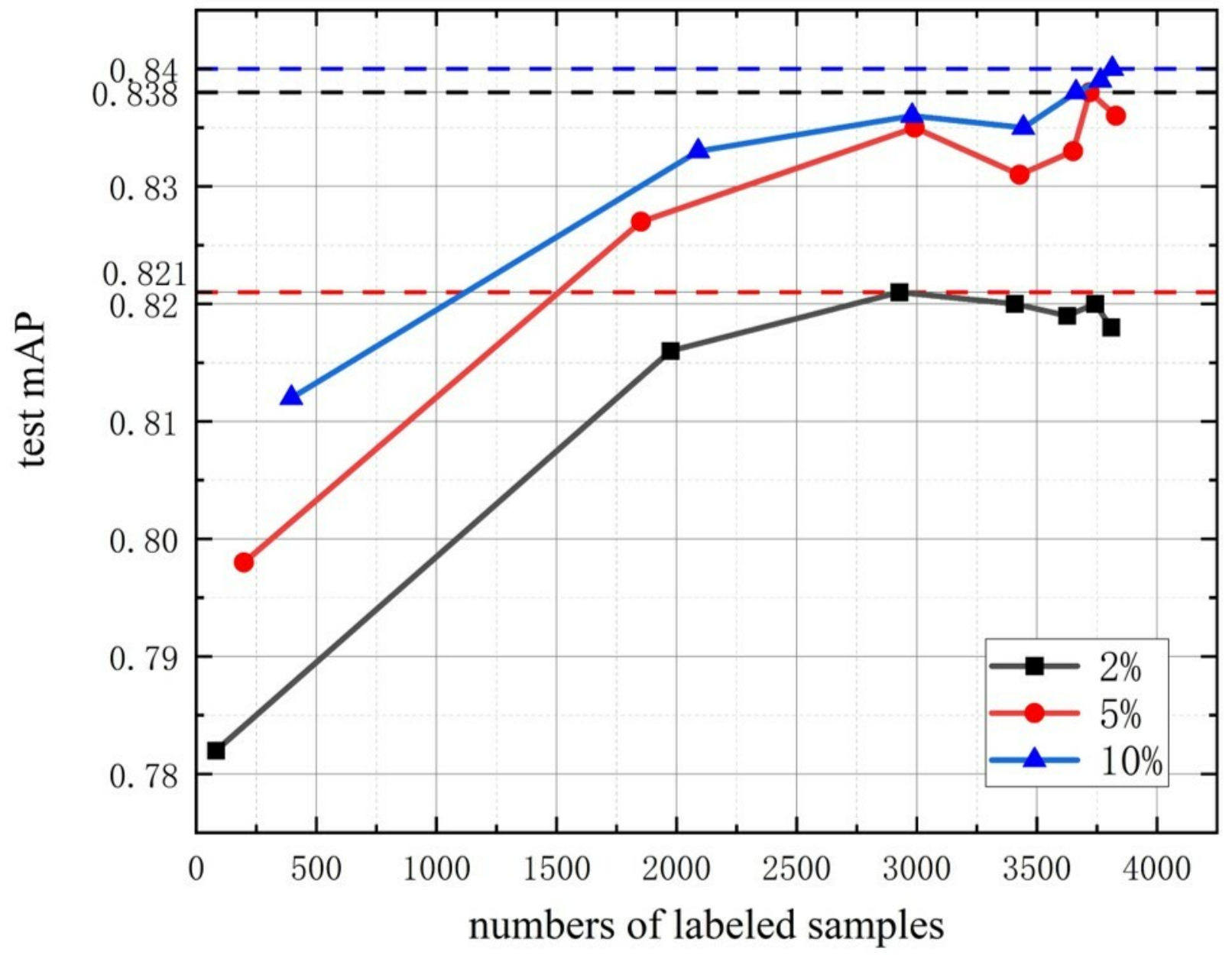}
    \caption{Effects of the number of key samples on Gram-SLD.}
    %\label{fig:myphoto}
\end{figure}
%%%%%%%%%%%%%%%%%%%%%%%% Effects of the number of self-labeling samples %%%%%%%%%%%%%%%%%%%%%%%%%
\subsubsection{Effects of the number of self-labeling samples}
In the self-labeling strategy of Subsection 3.3, the number of self-labeled samples in each iteration of co-training depends on parameter $\beta$. To investigate the effects of the number of self-labeled samples to be updated into the training pool in each iteration on the final detection accuracy and training time, we train Gram-SLD for $\beta$ =1.5, 1.0 and 0.5 on GMU Kitchen Dataset with RetinaNet as the original detection network. In addition, 199 labeled samples (5\% of all) are used as initial key samples, while the remaining 3652 are initially unlabeled. The results are summarized in Table 5. As one can see, when the ratio of self-labeled samples selected into the training pool is small (i.e., when $\beta$ is large), the algorithm can achieve high detection accuracy but requires more training time. On the other hand, when the ratio is large, training time can be saved but at the cost of lower detection accuracy. To balance the time cost and detection accuracy, we choose $\beta$ = 1.0, which leads to a sufficiently high accuracy (only 0.1\% difference to $\beta$ is 1.5) with reasonable training time.
\begin{table}[h]
    \centering
    \caption{Effect of $\beta$ on mAP and training times.}
    \begin{tabular}{|c|c|c|c|}
        \hline
        $\beta$ & samples updated & Training iterations & mAP\\
        \hline
        1.5 & 25\% & 15 & 83.9\%\\
        \hline
        1.0 & 50\% & 7 & 83.8\%\\
        \hline
        0.5 & 75\% & 5 & 83.3\%\\
        \hline
    \end{tabular}
\end{table}
%%%%%%%%%%%%%%%%%%%%%%%% Ablation study on gram loss %%%%%%%%%%%%%%%%%%%%%%%%%
\subsubsection{Ablation study on gram loss}
In Subsection 3.2, we define the total loss of the Gram-SLD in equation (1) by introducing a gram matrix-based loss term. In this subsection, we carry out an ablation study of Gram-SLD to study the effects of the gram matrix-based loss term on the independence of the two detectors in co-training. In this experiment, we use GMU Kitchen Dataset and apply the key sample selection strategy to select 199 samples (5\%) as key samples with the remaining unlabeled ones to be used in co-training.

The experimental results are shown in Table 5. As one can see, co-training without the gram loss term achieves the final mAP of 81.5\% with RetinaNet and 76.1\% with Faster R-CNN. On the other hand, adding the gram loss term manages to improve the accuracy to 83.8\% and 79.2\%, respectively. The reason is that gram loss makes two detectors less correlated, so that they can play the role of mutual complement and mutual correction. In this case, even if the accuracy of one detector decreases due to inaccurate annotation in a certain training iteration, the other detector still has the potential to correct the labeling result thanks to the independence of the two detectors, and, thus, leading to improved detection accuracy. In the absence of gram loss, however, once the result deviation of a certain round of labeling is large, the accuracy of the detector is likely to reach plateau or even decline continuously, due to the lack of a correction check mechanism.
\begin{table}[h]
    \centering
    \caption{Ablation experiment on gram loss.}
    \begin{tabular}{|c|c|c|}
        \hline
        model & Gram-SLD (proposed) & Learning w/o gram loss\\
        \hline
        RetinaNet & 83.8\% & 81.5\%\\
        \hline
        Faster R-CNN & 79.2\% & 76.1\%\\
        \hline
    \end{tabular}
\end{table}

%%%%%%%%%%%%%%%%%%%%%%%%%%%%%%%%%%%%%%%%%%%%%%%%%%%% 4.2 Comparison Experiment %%%%%%%%%%%%%%%%%%%%%%%%%%%%%
\subsection{Comparison Experiment}

After verifying the efficacy of the proposed Gram-SLD framework (Subsection 4.1) and studying the effects of various parameters on the algorithm’s detection performance (Subsection 4.2), in this subsection, we compare the proposed Gram-SLD method with two representative data-expanding training methods in the literature, Cut, paste and learn and Synthesizing training data.
%%%%%%%%%%%%%%%%%%%%%%%% Cut, paste and learn %%%%%%%%%%%%%%%%%%%%%%%%%
\subsubsection{Cut, paste and learn}
Cut, paste and learn generates 5759 synthetic samples using 6 instances in BigBIRD as objects and UW Scenes Dataset images as background. This method uses all synthesized samples and 10\% of the AVD training set (following the same training/test split described in Subsection 4.1.2), on which a Faster R-CNN with VGG16 is trained and achieves 78.7\% mAP on the AVD test set. Instead of blurring the whole patch as in the original Cut, paste and learn work, Rui Wang et al. have also performed data synthesis by only edge-smoothing along the object contours. In this work, the pruned Alexnet achieves 69.3\% mAP on the original GMU Kitchen dataset, with extended data, it can achieve 79.3\% mAP.

As a comparison, we manually annotate 10\% of the AVD training dataset and apply Gram-SLD with Faster R-CNN as the original detection network. As a result, the proposed Gram-SLD only achieves 78.5\% mAP through 8 iterations of co-training. Therefore, Cut, paste and learn is slightly more accurate than Gram-SLD in terms of mAP. However, it should be noted that both the Cut, paste and learn algorithm and our synthesis work require many multi-view masks of the object to generate the synthetic data. In these experiments, the multi-view masks of the target object can be directly obtained from the BigBIRD database (each object is shot at a vertical angle of every 18 degrees, and a horizontal angle of every 3 degrees, that is, each object has 600 multi-view masks). In real-life detection tasks, obtaining this many multi-view masks beforehand for the objects of interest can be extremely challenging. On the other hand, Gram-SLD only requires a certain number of unlabeled pictures that contain the objects of interest, has no strict restrictions on the views of the objects, and still performs on a similar level with Cut, paste and learn. Therefore, our proposed Gram-SLD has greater utility in practical applications.
%%%%%%%%%%%%%%%%%%%%%%%% Synthesizing training data %%%%%%%%%%%%%%%%%%%%%%%%%
\subsubsection{Synthesizing training data}
The synthesizing training data method generates about 20,000 synthetic samples using the same instances in BigBIRD as objects and NYU Depth v2 as background dataset. This method uses all synthetic samples and 10\% of the GMU training set as training data. Training a Faster R-CNN with VGG16 and achieves 79.2\% mAP on the GMU test set, while using 100\% samples of GMU training set can increase mAP to 82.5\% (i.e., an improvement of DIFF = 82.5\% – 79.2\% = 3.3\%). In the case of Gram-SLD, we first manually annotate 10\% of the GMU training use treats the remaining 90\% unlabeled samples via co-training. As a result, we achieve 79.5\% mAP, while having 100\% samples in GMU training set manually labeled can increase mAP to 80.8\% (i.e., an improvement of 1.3\%). The results are shown in Table 6. Note that the data of the training data synthesizing method is directly quoted from (\cite{RN136}) to eliminate the effects of different configurations during reproduction of the method. Apparently, compared to the method in (\cite{RN136}), Gram-SLD is more capable of exploiting the potential of the learning network with only a smaller number of labeled samples.
\begin{table}[h]
    \centering
    \caption{Comparison experiment with synthesizing training data method.}
     \resizebox{\textwidth}{20mm}{
    \begin{tabular}{|c|c|c|c|c|c|}
        \hline
        \multirow{2}{*}{method} & \multicolumn{2}{|c|}{Experiment 1} & \multicolumn{2}{c|}{Experiment 2} & \multirow{2}{*}{DIFF}\\
        \cline{2-5}
        & Training data & mAP & Training data & mAP & \\
        \hline
        \makecell[c]{Synthesizing \\ training data} & 
        \makecell[c]{10\% (labeled) GMU \\ training set + 20,000 \\ synthetic samples} &
        79.2\% &
        \makecell[c]{100\% (labeled) GMU \\ training set}& 
        82.5\% &
        3.3\%\\
        \hline
        \makecell[c]{Gram-SLD} & 
        \makecell[c]{10\% (labeled) + 90\% \\ (unlabeled) from GMU \\ training set} &
        79.5\% &
        \makecell[c]{100\% (labeled) GMU \\ training set}& 
        80.8\% &
        1.3\%\\
        \hline
    \end{tabular}
    }
\end{table}

%%%%%%%%%%%%%%%%%%%%%%%%%%%%%%%%%%%%%%%%%%%%%%% Conclusion %%%%%%%%%%%%%%%%%%%%%%%%%%%%%%%%%%%%%%%%%%%
\section{Conclusion}

%%%%%%%%%%%%%%%%%%%%%%%%%%%%%%%%%%%%%%%%%%%%%%%%% i am here??????????????????? %%%%%%%%%%%%%%%%%%%%%%%%%%%%%%%%%%%%%%%%%%%%%%
This paper proposes a semi-supervised learning framework for instance object detection via gram loss-guided co-training, which can reduce the dependence of the two detectors in co-training with the same input data. The main contributions of this work are:
\begin{enumerate}[(1)]
	\item  Developing a set of co-training mechanisms including network structure construction, initial key sample selection strategy, self-labeling strategy.
	\item Introducing a novel approach to alleviating the dependence of the detectors by introducing a gram matrix-based loss function on the feature maps of the detectors.
	\item  Developing a set of co-training mechanisms including network structure construction, initial key sample selection strategy, self-labeling strategy.
	\item Providing guidelines in selecting values of critical parameters of the algorithm in actual implementation. Experiments on different datasets and with detection networks demonstrate that, with only manually labeling 5\% of the samples in the training data is necessary and sufficient for the proposed Gram-SLD approach to achieve high quality performance compared with those state-of-the-art instance object detection methods, which may require significantly additional efforts in preparing extra data in the training dataset.
\end{enumerate}

 Thus, we claim that the proposed Gram-SLD is an effective and efficient approach to instance object detection.

%%%%%%%%%%%%%%%%%%%%%%%%%%%%%%%%%% 致谢 %%%%%%%%%%%%%%%%%%%%%%%%%%%%%%%%%%%%%%%%%%%
% % Acknowledgements should go at the end, before appendices and references

% \acks{We would like to acknowledge support for this project
% from the National Science Foundation (NSF grant IIS-9988642)
% and the Multidisciplinary Research Program of the Department
% of Defense (MURI N00014-00-1-0637). }
\section*{Acknowledgement}

This work was supported by the National Natural Science 
Foundation of China under Grant 61673039. The authors sincerely want to 
express heartfelt thanks to the editors and anonymous reviewers for the 
valuable comments and suggestions. We are particularly grateful to Jingwen 
Xu for the meaningful discussion and her work on the exploration of solving 
the problem of independence of viewpoint in co-training.

% Manual newpage inserted to improve layout of sample file - not
% needed in general before appendices/bibliography.

\newpage

% \appendix
% \section*{Appendix A.}
% \label{app:theorem}

% Note: in this sample, the section number is hard-coded in. Following
% proper LaTeX conventions, it should properly be coded as a reference:

% \noindent
% {\bf Theorem} {\it Let $u,v,w$ be discrete variables such that $v, w$ do
% not co-occur with $u$ (i.e., $u\neq0\;\Rightarrow \;v=w=0$ in a given
% dataset $\dataset$). Let $N_{v0},N_{w0}$ be the number of data points for
% which $v=0, w=0$ respectively, and let $I_{uv},I_{uw}$ be the
% respective empirical mutual information values based on the sample
% $\dataset$. Then
% \[
% 	N_{v0} \;>\; N_{w0}\;\;\Rightarrow\;\;I_{uv} \;\leq\;I_{uw}
% \]
% with equality only if $u$ is identically 0.} \hfill\BlackBox

% \noindent
% {\bf Proof}. We use the notation:\part{title}
% \[
% P_v(i) \;=\;\frac{N_v^i}{N},\;\;\;i \neq 0;\;\;\;
% P_{v0}\;\equiv\;P_v(0)\; = \;1 - \sum_{i\neq 0}P_v(i).
% \]
% These values represent the (empirical) probabilities of $v$
% taking value $i\neq 0$ and 0 respectively.  Entropies will be denoted
% by $H$. We aim to show that $\fracpartial{I_{uv}}{P_{v0}} < 0$....\\

% {\noindent \em Remainder omitted in this sample. See http://www.jmlr.org/papers/ for full paper.}

% 引入参考文献文件sample.bib
\bibliography{mybibfile}

\begin{thebibliography}{10}
\expandafter\ifx\csname url\endcsname\relax
  \def\url#1{\texttt{#1}}\fi
\expandafter\ifx\csname urlprefix\endcsname\relax\def\urlprefix{URL }\fi
\expandafter\ifx\csname href\endcsname\relax
  \def\href#1#2{#2} \def\path#1{#1}\fi
\bibitem{RN13}
G.~Georgakis, M.~A. Reza, A.~Mousavian, P.-H. Le, J.~Košecká, Multiview rgb-d
dataset for object instance detection, in: 2016 Fourth International
Conference on 3D Vision (3DV), IEEE, 2016, pp. 426--434.

\bibitem{RN11}
P.~Ammirato, P.~Poirson, E.~Park, J.~Košecká, A.~C. Berg, A dataset for
developing and benchmarking active vision, in: 2017 IEEE International
Conference on Robotics and Automation (ICRA), IEEE, 2017, pp. 1378--1385.

\bibitem{dataest}
R.~Wang, C.~Wu, \href{https://doip.buaa.edu.cn/info/1092/1103.htm}{Instance
object detection dataset [eb/ol]} (2021).
\newline\urlprefix\url{https://doip.buaa.edu.cn/info/1092/1103.htm}

\bibitem{RN8}
A.~Tejani, D.~Tang, R.~Kouskouridas, T.-K. Kim, Latent-class hough forests for
3d object detection and pose estimation, in: European Conference on Computer
Vision, Springer, 2014, pp. 462--477.

\bibitem{RN159}
R.~Wang, Y.~Liang, J.~Xu, Z.~H. He, Cascading classifier with discriminative
multi-features for a specific 3d object real-time detection, The Visual
Computer 35 (2018) 399--414.

\bibitem{2020Recent}
X.~Wu, D.~Sahoo, S.~Hoi, Recent advances in deep learning for object detection,
Neurocomputing 396 (2020).

\bibitem{2020Deep}
X.~Zeng, L.~Wen, B.~Liu, X.~Qi, Deep learning for ultrasound image caption
generation based on object detection, Neurocomputing 392 (2020) 132--141.

\bibitem{RN45}
R.~Girshick, J.~Donahue, T.~Darrell, J.~Malik, Rich feature hierarchies for
accurate object detection and semantic segmentation, in: Proceedings of the
IEEE conference on computer vision and pattern recognition, 2014, pp.
580--587.

\bibitem{RN153}
S.~Ren, K.~He, R.~B. Girshick, J.~Sun, Faster r-cnn: Towards real-time object
detection with region proposal networks, IEEE Transactions on Pattern
Analysis and Machine Intelligence 39 (2015) 1137--1149.

\bibitem{RN130}
J.~Dai, Y.~Li, K.~He, J.~Sun, R-fcn: Object detection via region-based fully
convolutional networks, ArXiv abs/1605.06409 (2016).

\bibitem{ZHANG2020180}
J.~Zhang, X.~Wu, S.~C. Hoi, J.~Zhu,
\href{https://www.sciencedirect.com/science/article/pii/S0925231219315309}{Feature
agglomeration networks for single stage face detection}, Neurocomputing 380
(2020) 180--189.
\newblock \href {https://doi.org/https://doi.org/10.1016/j.neucom.2019.10.087}
{\path{doi:https://doi.org/10.1016/j.neucom.2019.10.087}}.
\newline\urlprefix\url{https://www.sciencedirect.com/science/article/pii/S0925231219315309}

\bibitem{RN32}
D.~Dwibedi, I.~Misra, M.~Hebert, Cut, paste and learn: Surprisingly easy
synthesis for instance detection, in: Proceedings of the IEEE International
Conference on Computer Vision, 2017, pp. 1301--1310.

\bibitem{RN136}
G.~Georgakis, A.~Mousavian, A.~Berg, J.~Kosecka, Synthesizing training data for
object detection in indoor scenes, ArXiv abs/1702.07836 (2017).

\bibitem{RN175}
X.~Zhu, Semi-supervised learning literature survey, University of
Wisconsin-Madison (2008).

\bibitem{RN58}
T.~Joachims, Transductive inference for text classification using support
vector machines, in: Icml, Vol.~99, 1999, pp. 200--209.

\bibitem{RN120}
N.~Abe, H.~Mamitsuka, Query learning strategies using boosting and bagging, in:
ICML, 1998.

\bibitem{RN165}
Z.~Zhou, Semi-supervised learning by disagreement, in: GrC, 2010.

\bibitem{RN48}
A.~Blum, T.~Mitchell, Combining labeled and unlabeled data with co-training,
in: Proceedings of the eleventh annual conference on Computational learning
theory, 1998, pp. 92--100.

\bibitem{WU20201}
X.~Wu, D.~Sahoo, D.~Zhang, J.~Zhu, S.~C. Hoi,
\href{https://www.sciencedirect.com/science/article/pii/S0925231220303635}{Single-shot
bidirectional pyramid networks for high-quality object detection},
Neurocomputing 401 (2020) 1--9.
\newblock \href {https://doi.org/https://doi.org/10.1016/j.neucom.2020.02.116}
{\path{doi:https://doi.org/10.1016/j.neucom.2020.02.116}}.
\newline\urlprefix\url{https://www.sciencedirect.com/science/article/pii/S0925231220303635}

\bibitem{RN36}
D.~G. Lowe, Object recognition from local scale-invariant features, in:
Proceedings of the seventh IEEE international conference on computer vision,
Vol.~2, Ieee, 1999, pp. 1150--1157.

\bibitem{RN35}
H.~Bay, T.~Tuytelaars, L.~Van~Gool, Surf: Speeded up robust features, in:
European conference on computer vision, Springer, 2006, pp. 404--417.

\bibitem{RN33}
J.~Redmon, S.~Divvala, R.~Girshick, A.~Farhadi, You only look once: Unified,
real-time object detection, in: Proceedings of the IEEE conference on
computer vision and pattern recognition, 2016, pp. 779--788.

\bibitem{RN40}
W.~Liu, D.~Anguelov, D.~Erhan, C.~Szegedy, S.~Reed, C.-Y. Fu, A.~C. Berg, Ssd:
Single shot multibox detector, in: European conference on computer vision,
Springer, 2016, pp. 21--37.

\bibitem{RN34}
T.-Y. Lin, P.~Goyal, R.~Girshick, K.~He, P.~Dollár, Focal loss for dense
object detection, in: Proceedings of the IEEE international conference on
computer vision, 2017, pp. 2980--2988.

\bibitem{RN160}
R.~Wang, J.~Xu, T.~Han, Object instance detection with pruned alexnet and
extended training data, Signal Process. Image Commun. 70 (2019) 145--156.

\bibitem{RN90}
R.~Qin, W.~Rui, Generative deep deconvolutional neural network for increasing
and diversifying training data, in: 2018 IEEE International Conference on
Imaging Systems and Techniques (IST), 2018.

\bibitem{RN134}
M.~Fischler, R.~Bolles, Random sample consensus: a paradigm for model fitting
with applications to image analysis and automated cartography, Commun. ACM 24
(1981) 381--395.

\bibitem{RN155}
B.~M. Shahshahani, D.~Landgrebe, The effect of unlabeled samples in reducing
the small sample size problem and mitigating the hughes phenomenon, IEEE
Trans. Geosci. Remote. Sens. 32 (1994) 1087--1095.

\bibitem{RN176}
D.~J. Miller, H.~S. Uyar, A mixture of experts classifier with learning based
on both labelled and unlabelled data, in: NIPS, 1997.

\bibitem{RN149}
K.~Nigam, A.~McCallum, S.~Thrun, T.~M. Mitchell, Text classification from
labeled and unlabeled documents using em, Machine Learning 39 (2004)
103--134.

\bibitem{RN128}
A.~Blum, S.~Chawla, Learning from labeled and unlabeled data using graph
mincuts, in: ICML, 2001.

\bibitem{RN77}
X.~Zhu, Z.~Ghahramani, J.~D. Lafferty, Semi-supervised learning using gaussian
fields and harmonic functions, in: Machine Learning, Proceedings of the
Twentieth International Conference (ICML 2003), August 21-24, 2003,
Washington, DC, USA, 2003.

\bibitem{RN126}
M.~Belkin, P.~Niyogi, Semi-supervised learning on riemannian manifolds, Machine
Learning 56 (2004) 209--239.

\bibitem{RN151}
D.~Pierce, C.~Cardie, Limitations of co-training for natural language learning
from large datasets, in: EMNLP, 2011.

\bibitem{RN60}
A.~Sarkar, Applying co-training methods to statistical parsing, in: Second
Meeting of the North American Chapter of the Association for Computational
Linguistics, 2001.

\bibitem{RN61}
M.~Steedman, M.~Osborne, A.~Sarkar, S.~Clark, R.~Hwa, J.~Hockenmaier,
P.~Ruhlen, S.~Baker, J.~Crim, Bootstrapping statistical parsers from small
datasets, in: 10th Conference of the European Chapter of the Association for
Computational Linguistics, 2003.

\bibitem{RN62}
R.~Hwa, M.~Osborne, A.~Sarkar, M.~Steedman, Corrected co-training for
statistical parsers, in: Working Notes of the ICML’03 Workshop on the
Continuum from Labeled to Unlabeled Data in Machine Learning and Data Mining,
Citeseer, 2003.

\bibitem{RN64}
Z.-H. Zhou, K.-J. Chen, Y.~Jiang, Exploiting unlabeled data in content-based
image retrieval, in: European Conference on Machine Learning, Springer, 2004,
pp. 525--536.

\bibitem{RN163}
Z.~Zhou, K.-J. Chen, H.~Dai, Enhancing relevance feedback in image retrieval
using unlabeled data, ACM Transactions on Information Systems (TOIS) 24
(2006) 219 -- 244.

\bibitem{RN49}
K.~Nigam, R.~Ghani, Analyzing the effectiveness and applicability of
co-training, in: Proceedings of the ninth international conference on
Information and knowledge management, 2000, pp. 86--93.

\bibitem{RN52}
S.~Goldman, Y.~Zhou, Enhancing supervised learning with unlabeled data, in:
ICML, Citeseer, 2000, pp. 327--334.

\bibitem{RN67}
Y.~Zhou, S.~Goldman, Democratic co-learning, in: 16th IEEE International
Conference on Tools with Artificial Intelligence, IEEE, 2004, pp. 594--602.

\bibitem{RN164}
Z.~Zhou, M.~Li, Tri-training: exploiting unlabeled data using three
classifiers, IEEE Transactions on Knowledge and Data Engineering 17 (2005)
1529--1541.

\bibitem{RN144}
M.~Li, Z.~Zhou, Improve computer-aided diagnosis with machine learning
techniques using undiagnosed samples, IEEE Transactions on Systems, Man, and
Cybernetics - Part A: Systems and Humans 37 (2007) 1088--1098.

\bibitem{RN68}
Z.~Zhou, M.~Li, Semi-supervised learning with co-training, in: Proceedings of
the 19th International Joint Conference on Artificial Intelligence, 2006, pp.
908--913.

\bibitem{RN69}
U.~Brefeld, T.~Gärtner, T.~Scheffer, S.~Wrobel, Efficient co-regularised least
squares regression, in: Proceedings of the 23rd international conference on
Machine learning, 2006, pp. 137--144.

\bibitem{RN125}
M.-F. Balcan, A.~Blum, K.~Yang, Co-training and expansion: Towards bridging
theory and practice, in: NIPS, 2005.

\bibitem{RN70}
W.~Wang, Z.-H. Zhou, Analyzing co-training style algorithms, in: European
conference on machine learning, Springer, 2007, pp. 454--465.

\bibitem{RN135}
L.~A. Gatys, A.~S. Ecker, M.~Bethge, A neural algorithm of artistic style,
ArXiv abs/1508.06576 (2015).

\bibitem{RN161}
J.~H. Ward, Hierarchical grouping to optimize an objective function, Journal of
the American Statistical Association 58 (1963) 236--244.

\bibitem{RN16}
J.~MacQueen, Some methods for classification and analysis of multivariate
observations, in: Proceedings of the fifth Berkeley symposium on mathematical
statistics and probability, Vol.~1, Oakland, CA, USA, 1967, pp. 281--297.

\bibitem{RN158}
J.~Wang, W.~Zhang, T.~Hua, T.~Wei, Unsupervised learning of topological phase
transitions using calinski-harabaz index, arXiv: Statistical Mechanics
(2020).

\bibitem{RN12}
A.~Singh, J.~Sha, K.~S. Narayan, T.~Achim, P.~Abbeel, Bigbird: A large-scale 3d
database of object instances, in: 2014 IEEE international conference on
robotics and automation (ICRA), IEEE, 2014, pp. 509--516.

\bibitem{RN122}
P.~Ammirato, C.-Y. Fu, M.~Shvets, J.~Kosecka, A.~Berg, Target driven instance
detection, ArXiv abs/1803.04610 (2018).
\end{thebibliography}

\end{document}